\documentclass{article}
\usepackage[latin9]{inputenc}
\usepackage{array}
\usepackage{textcomp}
\usepackage{multirow}
\usepackage{graphicx}
\usepackage{microtype}
\usepackage{amsmath}
\usepackage{booktabs}
\usepackage[unicode=true,
 bookmarks=false,
 breaklinks=false,pdfborder={0 0 1},backref=section,colorlinks=false]
 {hyperref}

\makeatletter

\providecommand{\tabularnewline}{\\}

\usepackage{subfigure}

\usepackage[accepted]{icml2019}

\icmltitlerunning{Unsupervised Label Noise Modeling and Loss Correction}

\makeatother

\begin{document}
\twocolumn[
\icmltitle{Unsupervised Label Noise Modeling and Loss Correction}
\icmlsetsymbol{equal}{*}
\begin{icmlauthorlist}
\icmlauthor{Eric Arazo}{equal,to}
\icmlauthor{Diego Ortego}{equal,to}
\icmlauthor{Paul Albert}{to}
\icmlauthor{Noel E. O'Connor}{to}
\icmlauthor{Kevin McGuinness}{to}
\end{icmlauthorlist}

\icmlaffiliation{to}{Insight Centre for Data Analytics, Dublin City University (DCU), Dublin, Ireland}
\icmlcorrespondingauthor{Eric Arazo}{eric.arazo@insight-centre.org}
\icmlcorrespondingauthor{Diego Ortego}{diego.ortego@insight-centre.org}

\icmlkeywords{Machine Learning, ICML}
\vskip 0.3in
] 



\printAffiliationsAndNotice{\icmlEqualContribution} 
\begin{abstract}

Despite being robust to small amounts of label noise, convolutional neural networks trained with stochastic gradient methods have been shown to easily fit random labels. 
When there are a mixture of correct and mislabelled targets, networks tend to fit the former before the latter. 
This suggests using a suitable two-component mixture model as an unsupervised generative model of sample loss values during training to allow online estimation of the probability that a sample is mislabelled.
Specifically, we propose a beta mixture to estimate this probability and correct the loss by relying on the network prediction (the so-called bootstrapping loss). 
We further adapt \textit{mixup} augmentation to drive our approach a step further. 
Experiments on CIFAR-10/100 and TinyImageNet demonstrate a robustness to label noise that substantially outperforms recent state-of-the-art. Source code is available at \url{https://git.io/fjsvE}.
\end{abstract}

\section{Introduction}

Convolutional Neural Networks (CNNs) have recently become the par
excellence base approach to deal with many computer vision tasks~\cite{2016_arXiv_Homography,2018_arXiv_descriptors,2018_CVPR_ImClassif,2016_CVPR_YOLO,2017_CVPR_PSPNet,2017_ICCV_EventCaptioning}.
Their widespread use is attributable to their capability to model
complex patterns~\cite{2018_ICML_L2ReweightNoise} when vast amounts
of labeled data are available. Obtaining such volumes of data, however,
is not trivial and usually involves an error prone automatic or a
manual labeling process~\cite{2018_ECCV_DevilNoiseFace,2018_CVPR_LabelNoiseSemSeg}.
These errors lead to \emph{noisy samples}: samples annotated
with incorrect or \emph{noisy labels}. As a result, dealing with label
noise is a common adverse scenario that requires attention to ensure
useful visual representations can be learnt~\cite{2018_ICML_MentorNet,2018_ECCV_DevilNoiseFace,2018_TIFS_NoiseFaceRecog,2018_TTGRS_NoisyLabelHyperIm,2018_CVPR_LabelNoiseSemSeg}.
\begin{figure}[t]
\begin{centering}
\vskip 0.2in 
\par\end{centering}
\centering{}\renewcommand\tabcolsep{1.25 pt} 
\begin{tabular}{cc}
\includegraphics[width=0.5\columnwidth]{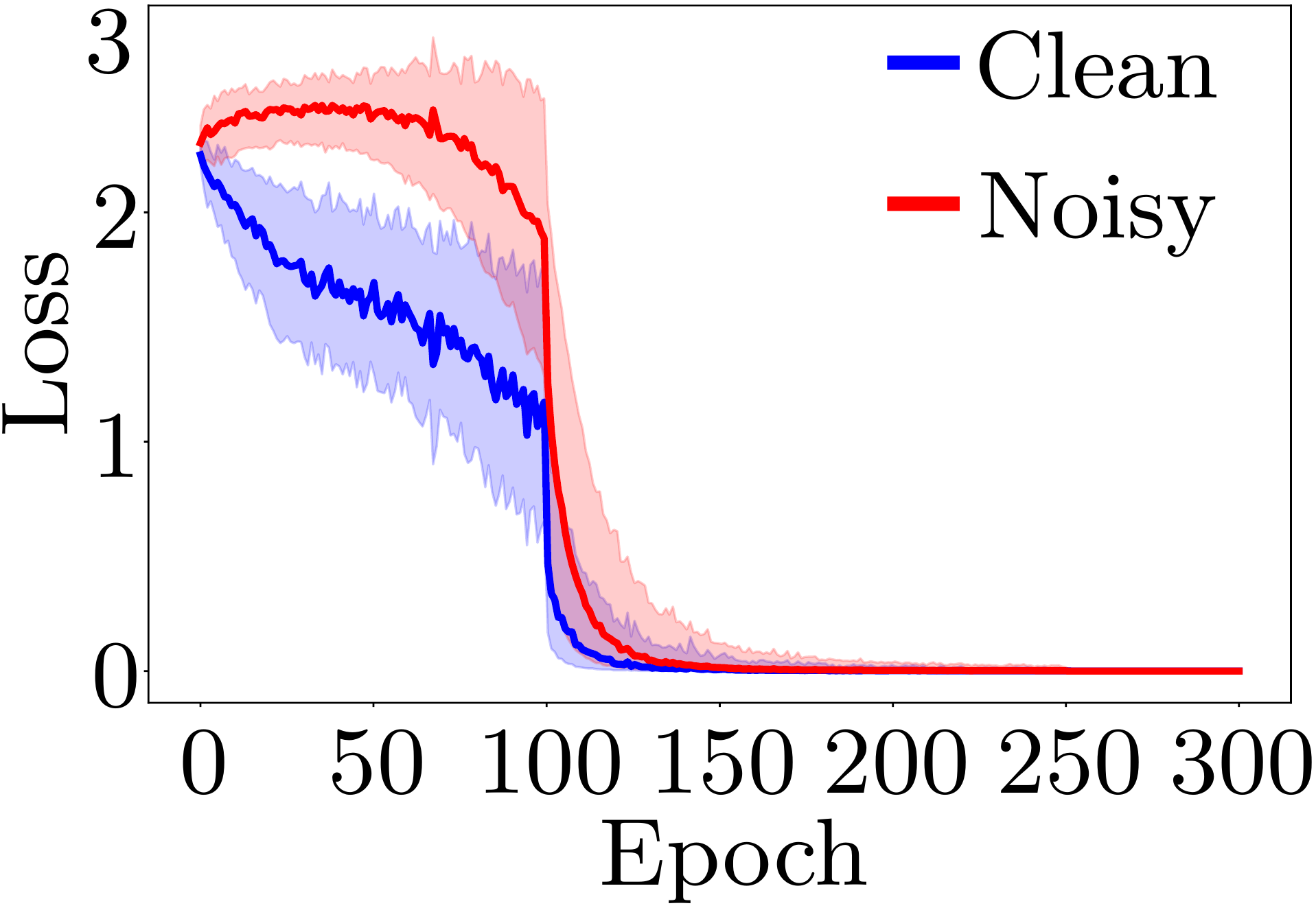} & \includegraphics[width=0.47\columnwidth]{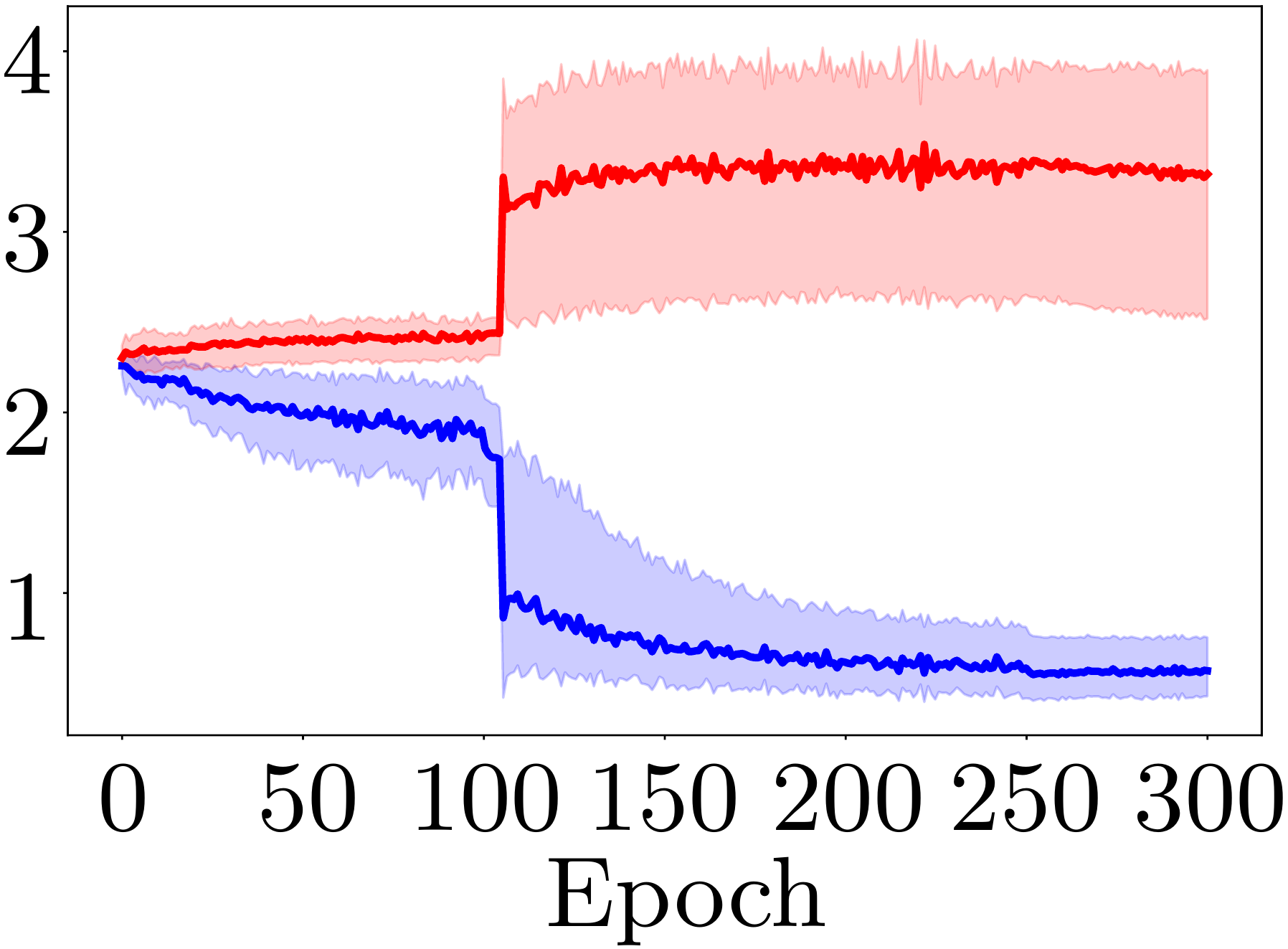}\tabularnewline
\end{tabular}\caption{\label{fig:Loss-during-training}Cross-entropy loss on CIFAR-10 under 80\% label noise for clean and noisy samples. Left: training with cross-entropy loss results in fitting the noisy labels. Right: using our proposed objective prevents fitting label noise while also learning from the noisy samples. The heavy lines represent the median losses and the shaded areas are the interquartile ranges.}
\vskip -0.2in 
\end{figure}
Automatically obtained noisy labels have previously been demonstrated
useful for learning visual representations~\cite{2017_CVPR_SelfSeg,2018_ICLR_Rotation};
however, a recent study on the generalization capabilities of deep
networks~\cite{2017_ICLR_Rethinking} demonstrates that noisy labels
are easily fit by CNNs, harming generalization. This overfitting also
arises in biases that networks encounter during training, e.g., when
a dataset contains class imbalances~\cite{2018_arxiv_BiasZisserman}.
However, before fitting label noise, CNNs fit the correctly labeled samples (\emph{clean samples)}
even under high-levels of corruption (Figure~\ref{fig:Loss-during-training}, left).

Existing literature on training with noisy labels focuses primarily
on loss correction approaches~\cite{2015_ICLR_Bootstrapping,2018_NIPS_GoldLoss,2018_ICML_MentorNet}.
A well-known approach is the bootstrapping loss~\cite{2015_ICLR_Bootstrapping},
which introduces a perceptual consistency term in the learning objective
that assigns a weight to the current network prediction to compensate
for the erroneous guiding of noisy samples. Other approaches modify
class probabilities~\cite{2017_CVPR_ForwardLoss,2018_NIPS_GoldLoss}
by estimating the noise associated with each class, thus computing
a loss that guides the training process towards the correct classes.
Still other approaches use curriculum learning to formulate a robust
learning procedure~\cite{2018_ICML_MentorNet,2018_ICML_L2ReweightNoise}.
Curriculum learning~\cite{2009_ICML_CurrLearn} is based on the idea
that ordering training examples in a meaningful (e.g. easy to hard)
sequence might improve convergence and generalization. In the noisy
label scenario, easy (hard) concepts are associated with clean (noisy)
samples by re-weighting the loss for noisy samples so that they contribute
less. Discarding noisy samples, however, potentially removes useful
information about the data distribution. \cite{2018_CVPR_IterativeNoise}
overcome this problem by introducing a similarity learning strategy
that pulls representations of noisy samples away from clean ones. Finally,
\emph{mixup} data augmentation~\cite{2018_ICLR_mixup} has recently
demonstrated outstanding robustness against label noise without explicitly
modeling it.

In light of these recent advances, this paper proposes a robust training
procedure that avoids fitting noisy labels even under high levels
of corruption (Figure~\ref{fig:Loss-during-training}, right), while
using noisy samples for learning visual representations that achieve
a high classification accuracy. Contrary to most successful recent
approaches that assume the existence of a known set of clean data~\cite{2018_ICML_L2ReweightNoise,2018_NIPS_GoldLoss},
we propose an unsupervised model of label noise based exclusively
on the loss on each sample. We argue that clean and noisy samples
can be modeled by fitting a two-component (clean-noisy) beta mixture
model (BMM) on the loss values. The posterior probabilities under
the model are then used to implement a dynamically weighted bootstrapping
loss, robustly dealing with noisy samples without discarding them.
We provide experimental work demonstrating the strengths
of our approach, which lead us to substantially outperform the related
work. Our main contributions are as follows: 
\begin{enumerate}
\item A simple yet effective unsupervised noise label modeling based on
each sample loss. 
\item A loss correction approach that exploits the unsupervised label noise
model to correct each sample loss, thus preventing overfitting to
label noise. 
\item Pushing the state-of-the-art one step forward by combining our approach
with mixup data augmentation~\cite{2018_ICLR_mixup}.
\item Guiding mixup data augmentation to achieve convergence even under
extreme label noise.
\end{enumerate}

\section{Related work}

Recent efforts to deal with label noise address two scenarios~\cite{2018_CVPR_IterativeNoise}:
closed-set and open-set label noise. In the closed set scenario, the
set of possible labels $S$ is known and fixed. All samples, including
noisy ones, have their true label in this set. In the open set scenario,
the true label of a noisy sample $x_{i}$ may be outside $S$; i.e.
$x_{i}$ may be an out-of-distribution sample~\cite{2018_ICLR_OutOfDist}.
The remainder of this section briefly reviews related work in the
closed-set scenario considered in~\cite{2017_ICLR_Rethinking}, upon
which we base our approach.

Several types of noise can be studied in the closed-set scenario,
namely \emph{uniform} or \emph{non-uniform} random label noise. The
former is also known as symmetric label noise and implies ground-truth
labels flipped to a different class with uniform random probability.
Non-uniform or class-conditional label noise, on the other hand, has
different flipping probabilities for each class~\cite{2018_NIPS_GoldLoss}.
Previous research~\cite{2017_CVPR_ForwardLoss} suggests that uniform
label noise is more challenging than non-uniform.

A simple approach to dealing with label noise is to remove the corrupted
data. This is not only challenging because difficult samples may be
confused with noisy ones~\cite{2018_CVPR_IterativeNoise}, but also
implies not exploiting the noisy samples for representation learning.
It has, however, recently been demonstrated~\cite{2018_WACV_SemiSupNoise}
that it is useful to discard samples with a high probability of being
incorrectly labeled and still use these samples in a semi-supervised
setup.

Other approaches seek to relabel the noisy samples by modeling their
noise through directed graphical models \cite{2015_CVPR_GraphModelNoise},
Conditional Random Fields \cite{2017_NIPS}, or CNNs \cite{2017_CVPR_CRFrelabel}.
Unfortunately, to predict the true label, these approaches rely on the
assumption that a small set of clean samples is always available,
which limits their applicability. Tanaka et al.~\cite{2018_CVPR_JointOpt}
have, however, recently demonstrated that it is possible to do unsupervised sample relabeling using the network predictions to
predict hard or soft labels.

Loss correction approaches~\cite{2015_ICLR_Bootstrapping,2018_ICML_MentorNet,2017_CVPR_ForwardLoss,2018_ICLR_mixup}
modify either the loss directly, or the probabilities used to compute
it, to compensate for the incorrect guidance provided by the noisy
samples. \cite{2015_ICLR_Bootstrapping} extend the loss with
a perceptual term that introduces a certain reliance on the model prediction. Their approach is, however, limited in that the noise
label always affects the objective. \cite{2017_CVPR_ForwardLoss}
propose a backward method that weights the loss of each sample using
the inverse of a noise transition matrix $T$, which specifies the probability
of one label being flipped to another. \cite{2017_CVPR_ForwardLoss}
presents a forward method that, instead of operating directly on the
loss, goes back to the predicted probabilities to correct them by
multiplying by the $T$ matrix.  \cite{2018_NIPS_GoldLoss}
corrects the predicted probabilities using a corruption matrix computed
using a model trained on a clean set of samples and their prediction
on the corrupted data. Other approaches focus on re-weighting
the contribution of noisy samples on the loss. \cite{2018_ICML_MentorNet}
proposes an alternating minimization framework in which a mentor network
learns a curriculum (i.e. a weight for each sample) to guide a student
network that learns under label noise conditions. Similarly, \cite{2018_ECCV_CurrNet}
present a curriculum learning approach based on an unsupervised estimation
on data complexity through its distribution in a feature space that
benefits from training with both clean and noisy samples. 
\cite{2018_ICML_L2ReweightNoise} weights each sample in the loss
based on the gradient directions in training compared to those on
validation (i.e. in a clean set). Note that, as for relabeling approaches,
the assumption of clean data availability limits the application of
many of these approaches. Conversely, approaches like \cite{2018_CVPR_IterativeNoise}
do not rely on clean data by performing unsupervised noise label detection
to help re-weighting the loss, while not discarding noisy samples
that are exploited in a similarity learning framework to pull their
representations away from true samples of each class.

In contrast to the aforementioned literature, we propose to deal with
noisy labels using exclusively the training loss of each sample without
consulting any clean set. Specifically, we fit a two-component
beta mixture model to the training loss of each sample to model clean
and noisy samples. We use this unsupervised model to implement
a loss correction approach that benefits both from bootstrapping \cite{2015_ICLR_Bootstrapping}
and mixup data augmentation \cite{2018_ICLR_mixup} to deal with the
closed-set label noise scenario.

\section{Learning with label noise}

Image classification can be formulated as the problem of learning a model $h_{\theta}(x)$ from a set of training examples $\mathcal{D}=\left\{ \left(x_{i},y_{i}\right)\right\} _{i=1}^{N}$ with $y_{i}\in\left\{ 0,1\right\} ^{C}$ being the one-hot encoding ground-truth label corresponding to $x_{i}$.
In our case, $h_{\theta}$ is a CNN and $\theta$ represents the model parameters (weights and biases).
As we are considering classification under label noise, the label $y_{i}$ can be noisy (i.e. $x_{i}$ is a noisy sample).
The parameters $\theta$ are fit by optimizing a loss function, e.g. categorical cross-entropy:
\begin{equation}
\ell(\theta)=\sum_{i=1}^{N} \ell_i(\theta) = -\sum_{i=1}^{N} y_{i}^T\log\left(h_{\theta}(x_{i})\right),\label{eq:Cross-ent}
\end{equation}
where $h_{\theta}(x)$ are the softmax probabilities produced by the model and $\log(\cdot)$ is applied elementwise. 
The remainder of this section describes our noisy sample modeling technique and how to extend the loss in Eq.~\eqref{eq:Cross-ent} based on this model to handle label noise. 
For notational simplicity, we use $\ell_i(\theta)=\ell_i$ and $h_{\theta}(x_i)=h_i$ in the remainder of the paper.

\subsection{Label noise modeling\label{subsec:Label-noise-modeling}}

\begin{figure}[t]
\centering{}\includegraphics[width=1\columnwidth]{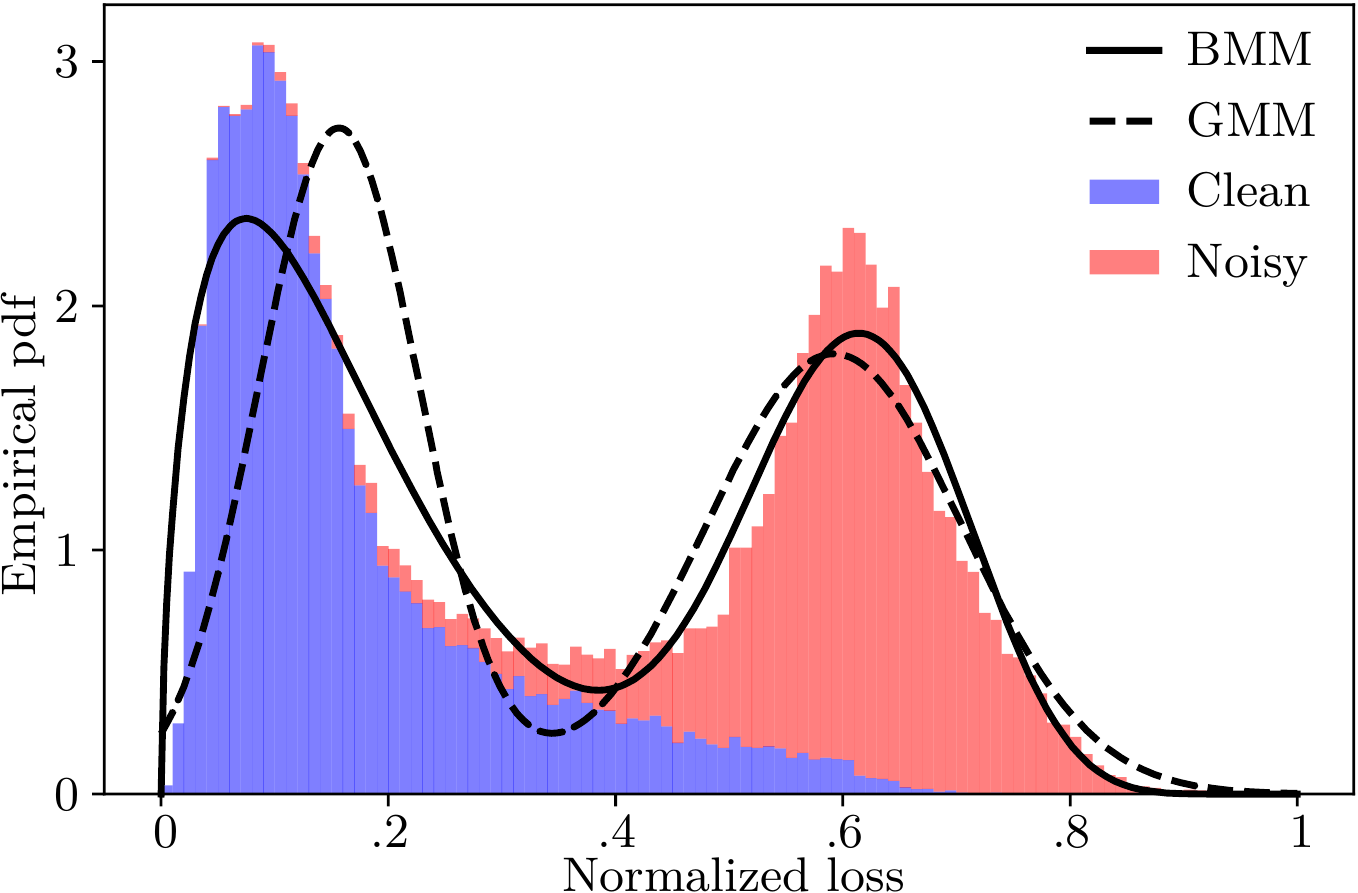}\caption{\label{fig:Normalized-histograms}Empirical PDF and estimated GMM and BMM models for 50\% label noise in CIFAR-10 after 10 epochs with standard cross-entropy loss and learning rate of 0.1 (remaining hyperparameters see in Subsection \ref{subsec:Datasets-and-implementation}). Clean and noisy samples are colored for illustrative purposes. The BMM model better fits the skew toward zero loss of the noisy samples.}
\end{figure}

We aim to identify the noisy samples in the dataset $\mathcal{D}$
so that we can implement a loss correction approach (see Subsections
\ref{subsec:DynamicBootstrapping} and \ref{subsec:JointMixBoot}).
Our essential observation is simple: random labels take longer to learn than
clean labels, meaning that noisy samples have higher loss during the early epochs
of training (see Figure \ref{fig:Loss-during-training}), allowing clean and noisy samples to be distinguished from the loss distribution alone (see Figure~\ref{fig:Normalized-histograms}). Modern CNNs trained 
with stochastic gradient methods typically do not fit the noisy examples until 
substantial progress has been made in fitting the clean ones. Therefore, one
can infer from the loss value if a sample is more likely to be clean or
noisy. We propose to use a mixture distribution model for this purpose.

Mixture models are a widely used unsupervised modeling technique
\cite{1999_GMM_BS,2006_PR_GMM_Imseg,2011_TPAMI_BMMestimation}, 
with the Gaussian Mixture Model (GMM) \cite{2006_PR_GMM_Imseg} being the most
popular. The probability density function (pdf) of a mixture model of
$K$ components on the loss $\ell$ is defined as: 
\begin{equation}
p\!\left(\ell\right)=\sum_{k=1}^{K}\lambda_{k}\,p\!\left(\ell\mid k\right),
\label{eq:mix}
\end{equation}
where $\lambda_{k}$ are the mixing coefficients for the convex combination
of each individual pdf $p\!\left(\ell\mid k\right)$. In our case,
we can fit a two components GMM (i.e. $K=2$ and $\ell\sim\mathcal{N}\left(\mu_{k},\sum_{k}\right)$)
to model the distribution of clean and noisy samples
(Figure~\ref{fig:Normalized-histograms}). Unfortunately, the Gaussian
is a poor approximation to the clean set
distribution, which exhibits high skew toward zero. The more flexible beta distribution \cite{2011_TPAMI_BMMestimation} allows modelling both symmetric and skewed distributions over $[0, 1]$; the beta mixture model (BMM) better approximates the loss distribution
for mixtures of clean and noisy samples (Figure \ref{fig:Normalized-histograms}). Empirically, we also found the BMM improves ROC-AUC for clean-noisy label classification over the GMM by around 5 points for 80\% label noise in CIFAR-10 when using the training objective in Section \ref{subsec:JointMixBoot} (see Appendix A).  The beta distribution over a (max) normalized loss $\ell\in\left[0,1\right]$
is defined to have pdf:

\begin{equation}
p\!\left(\ell\mid\alpha,\beta\right)=\frac{\Gamma\!\left(\alpha+\beta\right)}{\Gamma\!\left(\alpha\right)\Gamma\!\left(\beta\right)}\ell^{\alpha-1}\left(1-\ell\right)^{\beta-1},
\end{equation}
where $\alpha,\beta>0$ and $\Gamma\!\left(\cdot\right)$ is the Gamma
function, and the mixture pdf is given by substituting the above into Eq.~\eqref{eq:mix}.

We use an Expectation Maximization (EM) procedure to fit the BMM to
the observations. Specifically, we introduce latent variables $\ensuremath{\gamma_{k}(\ell)=p(k|\ell)}$
which are defined to be the posterior probability of the point $\ell$
having been generated by mixture component $\ensuremath{k}$. In the
E-step we fix the parameters $\ensuremath{\lambda_{k},\alpha_{k},\beta_{k}}$
and update the latent variables using Bayes rule: 
\begin{equation}
\gamma_{k}(\ell)=\frac{\lambda_{k}\,p\!\left(\ell\mid\alpha_{k},\beta_{k}\right)}{\sum_{j=1}^{K}\lambda_{j}\,p\!\left(\ell\mid\alpha_{j},\beta_{j}\right)}.
\end{equation}
Given fixed $\ensuremath{\gamma_{k}(\ell)}$, the M-step estimates
the distribution parameters $\ensuremath{\alpha_{k},\beta_{k}}$ using
a weighted version of the method of moments: 
\begin{equation}
\beta_{k}=\frac{\alpha_{k}\left(1-\bar{\ell}_{k}\right)}{\bar{\ell}_{k}},\;\alpha_{k}=\bar{\ell}_{k}\left(\frac{\bar{\ell}_{k}\left(1-\bar{\ell}_{k}\right)}{s_{k}^{2}}-1\right)
\end{equation}
with $\bar{\ell}_{k}$ being a weighted average of the losses $\left\{ \ell_{i}\right\} _{i=1}^{N}$
corresponding to each training sample $\left\{ x_{i}\right\} _{i=1}^{N}$,
and $s_{k}^{2}$ being a weighted variance estimate: 
\begin{equation}
\bar{\ell}_{k}=\frac{\sum_{i=1}^{N}\gamma_{k}(\ell_{i})\,\ell_{i}}{\sum_{i=1}^{N}\gamma_{k}(\ell_{i})},
\end{equation}
\begin{equation}
s_{k}^{2}=\frac{\sum_{i=1}^{N}\gamma_{k}(\ell_{i})\,\left(\ell_{i}-\bar{\ell}_{k}\right)^{2}}{\sum_{i=1}^{N}\gamma_{k}(\ell_{i})}.
\end{equation}
The updated mixing coefficients $\ensuremath{\lambda_{k}}$ are then
calculated in the usual way: 
\begin{equation}
\ensuremath{\lambda_{k}}=\frac{1}{N}\sum_{i=1}^{N}\gamma_{k}(\ell_{i}).
\end{equation}
The above E and M-steps are then iterated until convergence or a maximum
number of iterations (10 in our experiments) are reached. Note that
the above algorithm becomes numerically unstable when the observations
are very near zero and one. Our implementation simply sidesteps this
issue by bounding the observations in $\left[\epsilon,1-\epsilon\right]$
instead of {[}0, 1{]} ($\epsilon=10^{-4}$ in our experiments).

Finally, we obtain the probability of a sample being clean or noisy
through the posterior probability: 
\begin{equation}
p\!\left(k\mid\ell_{i}\right)=\frac{p\!\left(k\right)p\!\left(\ell_{i}\mid k\right)}{p\!\left(\ell_{i}\right)},
\end{equation}
where $k=0\left(1\right)$ denotes clean (noisy) classes.

Note that the loss used to estimate the mixture distribution is always the standard cross-entropy loss (Figure \ref{fig:Loss-during-training}) for all samples after every epoch. This not necessarily the loss used for training, which may contain a corrective component to deal with label noise.

\subsection{Noise model for label correction\label{subsec:DynamicBootstrapping}}

Carefully selecting a loss function to guide the learning process is of 
particular importance under label noise. Standard categorical cross-entropy
loss (Eq.~\eqref{eq:Cross-ent}) is ill-suited to the task as it encourages 
fitting label noise~\cite{2017_ICLR_Rethinking}.
The static hard bootstrapping loss proposed in \cite{2015_ICLR_Bootstrapping}
provides a mechanism to deal with label noise by adding a perceptual term to the standard
cross-entropy loss that helps to correct the training
objective: 
\begin{equation}
\ell_{B}=-\sum_{i=1}^{N}\left(\left(1-w_{i}\right)y_{i}+w_{i}z_{i}\right)^T\log\left(\mathit{h_{i}}\right),\label{eq:HardBoot}
\end{equation}
where $w_{i}$ weights the model prediction $z_{i}$ in the loss function.
\cite{2015_ICLR_Bootstrapping} use $w_{i}=0.2,\forall i$.
We refer to this approach as static hard bootstrapping. 
\cite{2015_ICLR_Bootstrapping} also proposed a static soft bootstrapping loss ($w_{i}=0.05,\forall i$)
that uses the predicted softmax probabilities $\mathit{h}_{i}$
instead of the class prediction $z_{i}$. Unfortunately, using a fixed
weight for all samples does not prevent fitting the noisy ones (Table \ref{tab:NmixDbootCIFAR} in Subsection \ref{subsec:Static-vs-Dynamic Boot})
and, more importantly, applying a small fixed weight $w_{i}$ to
the prediction (probabilities) $z_{i}$ ($\mathit{h}_{i}$) limits the correction of a hypothetical noisy label $y_{i}$.

We propose dynamic hard and soft bootstrapping losses by
using our noise model to individually weight each sample; i.e., $w_{i}$
is dynamically set to $p\!\left(k=1\mid\ell_{i}\right)$ and the BMM
model is estimated after each training epoch using the cross-entropy
loss for each sample $\ell_{i}$. Therefore, clean samples rely on
their ground-truth label $y_{i}$ ($1-w_{i}$ is large), while noisy
ones let their loss being dominated by their class prediction $z_{i}$
or their predicted probabilities $\mathit{h_{i}}$ ($w_{i}$ is large),
respectively, for hard and soft alternatives. Note that in mature
stages of training the CNN model should provide a good
estimation of the true class for noisy samples. Subsection
\ref{subsec:Static-vs-Dynamic Boot} compares static and dynamic 
bootstrapping, showing that dynamic bootstrapping gives superior results.

\subsection{Joint label correction and \emph{mixup} data augmentation\label{subsec:JointMixBoot}}

Recently \cite{2018_ICLR_mixup} proposed a data augmentation technique
named \emph{mixup} that exhibits strong robustness to label
noise. This technique trains on convex combinations of sample pairs ($x_{p}$ and $x_{q}$) and corresponding labels ($y_{p}$
and $y_{q}$): 
\begin{equation}
x=\delta x_{p}+(1-\delta)x_{q},
\end{equation}
\begin{equation}
\ell=\delta\ell_{p}+(1-\delta)\ell_{q},\label{eq:MixupLoss}
\end{equation}
where $\delta$ is randomly sampled from a beta distribution $\mathcal{B}e\left(\alpha,\beta\right)$,
with $\alpha=\beta$ set to high values when learning with label
noise so that $\delta$ tends to be close to 0.5. This combination regularizes the network to favor simple linear
behavior between training samples, which reduces oscillations
in regions far from them. Regarding label
noise, \emph{mixup} provides a mechanism to combine clean and noisy samples, computing a more representative loss to guide
the training process. Even when combining two noisy
samples the loss computed can still be useful as one of the noisy
samples may (by chance) contain the true label of the other one. As for preventing
overfitting to noisy samples, the fact that samples and their labels
are mixed favors learning structured data, while hindering learning
the unstructured noise.

\emph{Mixup} achieves robustness to label noise by 
appropriate combinations of training examples. Under high-levels of noise
 mixing samples that both have incorrect labels is prevalent, which reduces the effectiveness of the method. 
 We propose to fuse \emph{mixup} and our dynamic bootstrapping to implement a robust per-sample loss correction approach: 
\[
\ell^{*}=-\delta\left[\left(\left(1-w_{p}\right)y_{p}+w_{p}z_{p}\right)^T\log\left(\mathit{h}\right)\right]-
\]
\begin{equation}
\left(1-\delta\right)\left[\left(\left(1-w_{q}\right)y_{q}+w_{q}z_{q}\right)^T\log\left(\mathit{h}\right)\right],\label{eq:SMixDHardLoss}
\end{equation}
The loss $\ell^{*}$ defines the hard alternative, while the soft
one can be easily defined by replacing $z_{p}$ and $z_{q}$ by $h_{p}$
and $h_{q}$. These hard and soft losses exploit mixup's advantages
while correcting the labels through dynamic bootstrapping, i.e. the
weights $w_{p}$ and $w_{q}$ that control the confidence in the ground-truth
labels and network predictions are inferred from our unsupervised noise model:
$w_{p}=p\!\left(k=1\mid\ell_{p}\right)$ and $w_{q}=p\!\left(k=1\mid\ell_{q}\right)$.
We compute $h_{p}$, $z_{p}$, $h_{q}$ and $z_{q}$ by doing an extra forward pass, as it is not straightforward to obtain the predictions for samples $p$
and $q$ from the mixed probabilities $h$.

Ideally, the proposed loss $\ell^{*}$ would lead to a better model
by trusting in progressively better predictions during training. For high-levels of label noise, however, the network predictions are unreliable
and dynamic bootstrapping may not converge when combined with the
complex signal that \emph{mixup} provides. This is reasonable as under high
levels of noise most of the samples are guided by the network's prediction
in the bootstrapping loss, encouraging the network to predict the same
class to minimize the loss. We
apply the regularization term used in \cite{2018_CVPR_JointOpt},
which seeks preventing the assignment of all samples to a single class, to overcome this issue:
\begin{equation}
R=\sum_{c=1}^{C}p_{c}\log\left(\frac{p_{c}}{\overline{h}_{c}}\right),\label{eq:Regularization}
\end{equation}
where $p_{c}$ denotes the prior probability distribution for class
$c$ and $\overline{h}_{c}$ is the mean softmax probability of the
model for class $c$ across all samples in the dataset. Note that
we assume a uniform distribution for the prior probabilities (i.e.
$p_{c}=1/C$), while approximating $\overline{h}_{c}$ using mini-batches
as done in \cite{2018_CVPR_JointOpt}. We add the term $\eta R$ to $\ell^{*}$ (Eq.~\eqref{eq:SMixDHardLoss}) with $\eta$ being the regularization coefficient (set to one in all the experiments). Subsection~\ref{subsec:Joint-mixup-and- boot} presents the results of this
approach and Subsection~\ref{subsec:Comparison-against-related} demonstrates its superior performance in comparison to the state-of-the-art.

\section{Experiments}

\subsection{Datasets and implementation details\label{subsec:Datasets-and-implementation}}

We thoroughly validate our approach in two well-known image classification datasets:
CIFAR-10 and CIFAR-100. The former contains 10 classes, while the
latter has 100 classes. Both have 50K color images for training
and 10K for validation with resolution 32\texttimes 32. We use a PreAct
ResNet-18 \cite{2016_ECCV_PreActResNet} and train it using SGD and batch size of
128. We use two different schemes for the learning rate policy and number of epochs depending on whether \emph{mixup} is used (see Appendix B for further details). We further experiment on TinyImageNet (subset of ImageNet
\cite{2009_CVPR_ImageNet}) and Clothing1M \cite{2015_CVPR_GraphModelNoise}
 datasets to test the generality of our approach far from CIFAR data
(Subsection \ref{subsec:Additional datasets}). TinyImageNet contains
200 classes with 100K training images, 10K  validation, 
10K test with resolution $64\times64$, while Clothing1M contains
14 classes with 1M real-world noisy training samples and clean training subsets (47K), validation (14K) and test (10K).

We follow \cite{2017_ICLR_Rethinking,2018_ICLR_mixup,2018_CVPR_JointOpt}
criterion for label noise addition, which consists of randomly selecting
labels for a percentage of the training data using all possible labels
(i.e. the true label could be randomly maintained). Note that there
is another popular label noise criterion \cite{2018_ICML_MentorNet,2018_CVPR_IterativeNoise}
in which the true label is not selected when performing random labeling.
We also run our proposed approach under these conditions in Subsection
\ref{subsec:Comparison-against-related} for comparison.

\subsection{Static and dynamic loss correction\label{subsec:Static-vs-Dynamic Boot}}

\begin{table}[t]
\begin{centering}
\caption{\label{tab:NmixDbootCIFAR}Validation accuracy on CIFAR-10 for static bootstrapping
and the proposed dynamic bootstrapping. Key: CE (cross-entropy loss), ST
(static bootstrapping), DY (dynamic bootstrapping), S (soft), and H
(hard). Bold indicates best performance.}
\vskip 0.15in 
\par\end{centering}
\begin{centering}
\begin{small}%
\begin{tabular}{llrrrrr}
\toprule 
Alg./Noise level (\%) &  & 0 & 20 & 50 & 80\tabularnewline
\hline 
\multirow{2}{*}{CE} & Best & 93.8 & \textbf{89.7} & \textbf{84.8} & 67.8\tabularnewline
 & Last & 93.7 & 81.8 & 55.9 & 25.3\tabularnewline
\midrule 
\multirow{2}{*}{ST-S} & Best & \textbf{93.9} & \textbf{89.7} & 84.8 & 67.8\tabularnewline
 & Last & \textbf{93.9} & 81.7 & 55.9 & 24.8\tabularnewline
\midrule 
\multirow{2}{*}{ST-H} & Best & 93.8 & \textbf{89.7} & \textbf{84.8} & 68.0\tabularnewline
 & Last & 93.8 & 81.4 & 56.4 & 25.7\tabularnewline
\midrule 
\multirow{2}{*}{DY-S} & Best & 93.6 & \textbf{89.7} & \textbf{84.8} & 67.8\tabularnewline
 & Last & 93.4 & 83.3 & 57.0 & 27.8\tabularnewline
\midrule 
\multirow{2}{*}{DY-H} & Best & 93.3 & \textbf{89.7} & \textbf{84.8} & \textbf{71.7}\tabularnewline
 & Last & 92.9 & \textbf{83.4} & \textbf{65.0} & \textbf{64.2}\tabularnewline
\bottomrule 
\end{tabular}\end{small}
\par\end{centering}
\centering{}\vskip -0.1in 
\end{table}

\begin{table}[t]
\begin{centering}
\caption{\label{tab:JointPerformance}Validation accuracy on CIFAR-10 (top) and CIFAR-100
(bottom) for joint mixup and bootstrapping. Key: CE (cross-entropy), M (mixup), DYR (dynamic bootstrapping + regularization from
Eq. \ref{eq:Regularization}), S (soft), and H (hard). Bold indicates
best performance.}
\vskip 0.15in 
\par\end{centering}
\begin{centering}
\begin{small}%
\begin{tabular}{llrrrrr}
\toprule 
Alg./Noise level (\%) &  & 0 & 20 & 50 & 80\tabularnewline
\midrule 
\multirow{2}{*}{CE} & Best & 94.7 & 86.8 & 79.8 & 63.3\tabularnewline
 & Last & 94.6 & 82.9 & 58.4 & 26.3\tabularnewline
\midrule 
\multirow{2}{*}{M \cite{2018_ICLR_mixup}} & Best & \textbf{95.3} & \textbf{95.6} & 87.1 & 71.6\tabularnewline
 & Last & \textbf{95.2} & 92.3 & 77.6 & 46.7\tabularnewline
\midrule 
\multirow{2}{*}{M-DYR-S} & Best & 93.3 & 93.5 & 89.7 & 77.3\tabularnewline
 & Last & 93.0 & 93.1 & 89.3 & 74.1\tabularnewline
\midrule 
\multirow{2}{*}{M-DYR-H} & Best & 93.6 & 94.0 & \textbf{92.0} & \textbf{86.8}\tabularnewline
 & Last & 93.4 & \textbf{93.8} & \textbf{91.9} & \textbf{86.6}\tabularnewline
\bottomrule 
\toprule 
Alg./Noise level (\%) &  & 0 & 20 & 50 & 80\tabularnewline
\midrule 
\multirow{2}{*}{CE} & Best & \textbf{76.1} & 62.0 & 46.6 & 19.9\tabularnewline
 & Last & \textbf{75.9} & 62.0 & 37.7 & 8.9\tabularnewline
\midrule 
\multirow{2}{*}{M \cite{2018_ICLR_mixup}} & Best & 74.8 & 67.8 & 57.3 & 30.8\tabularnewline
 & Last & 74.4 & 66.0 & 46.6 & 17.6\tabularnewline
\midrule 
\multirow{2}{*}{M-DYR-S} & Best & 71.9 & 67.9 & \textbf{61.7} & 38.8\tabularnewline
 & Last & 67.4 & 67.5 & \textbf{58.9} & 34.0\tabularnewline
\midrule 
\multirow{2}{*}{M-DYR-H} & Best & 70.3 & \textbf{68.7} & \textbf{61.7} & \textbf{48.2}\tabularnewline
 & Last & 66.2 & \textbf{68.5} & 58.8 & \textbf{47.6}\tabularnewline
\bottomrule 
\end{tabular}\end{small}
\par\end{centering}
\centering{}\vskip -0.1in 
\end{table}
Table~\ref{tab:NmixDbootCIFAR} presents the results for static (ST) and
dynamic (DY) bootstrapping in CIFAR-10. Although ST achieves performance comparable to DY (except for 80\% noise where DY is much better), after the final epoch (last) the performance of DY outperforms ST. The improvements are particularly remarkable for 80\% of label noise (from 25.7\% of ST-H to 64.2 of DY-H). Comparing soft and hard alternatives: hard bootstrapping gives superior performance, which is consistent with the findings of the original paper \cite{2015_ICLR_Bootstrapping}. The overall results demonstrate that applying per-sample weights (DY) benefits training by allowing to fully correct noisy labels.

\subsection{Joint \emph{mixup} and dynamic loss correction\label{subsec:Joint-mixup-and- boot}}

\begin{figure}[t]
\centering{}\vskip 0.2in \renewcommand\tabcolsep{0.75 pt} 
\begin{tabular}{ccc}
\includegraphics[width=0.3275\columnwidth]{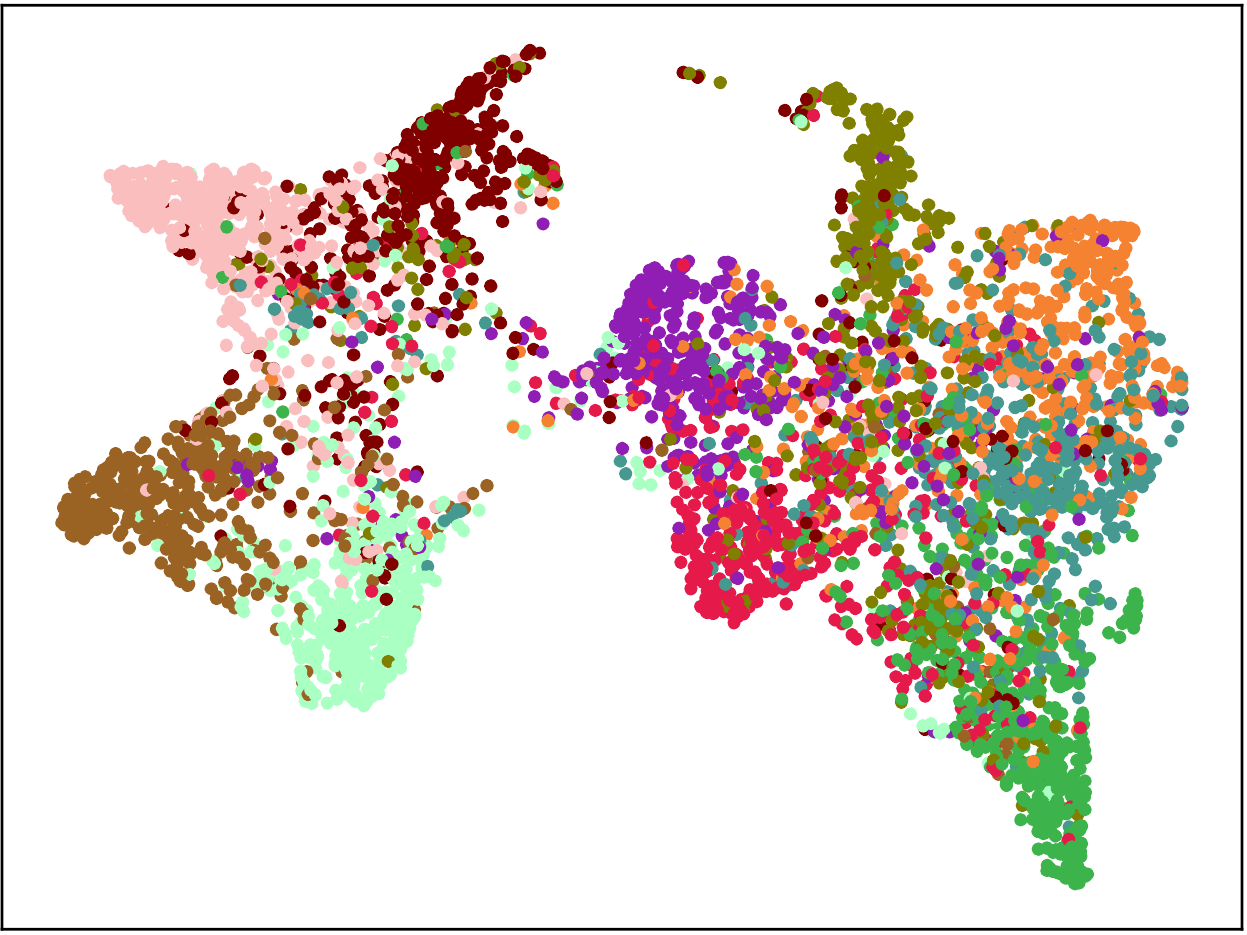} & \includegraphics[width=0.3275\columnwidth]{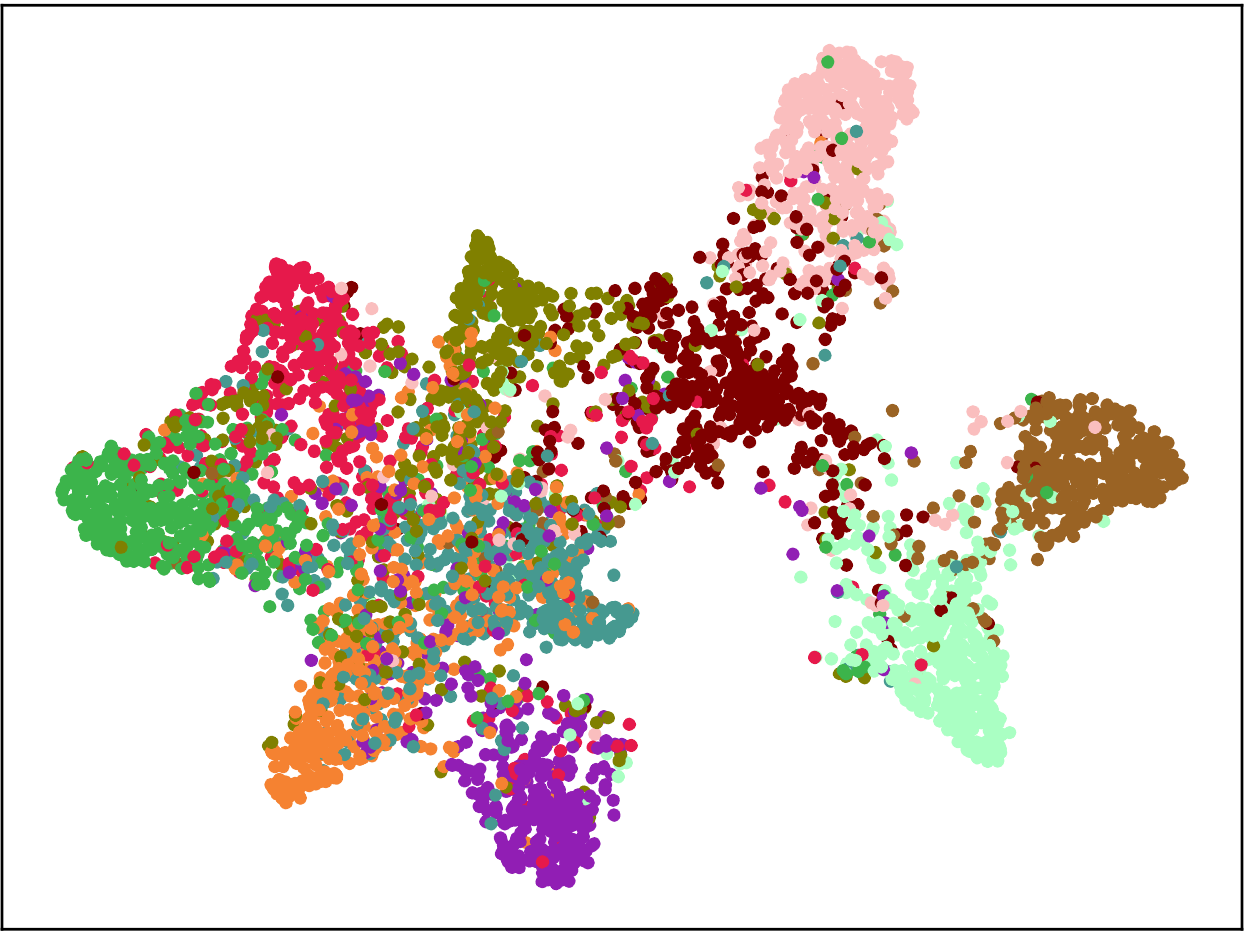} & \includegraphics[width=0.3275\columnwidth]{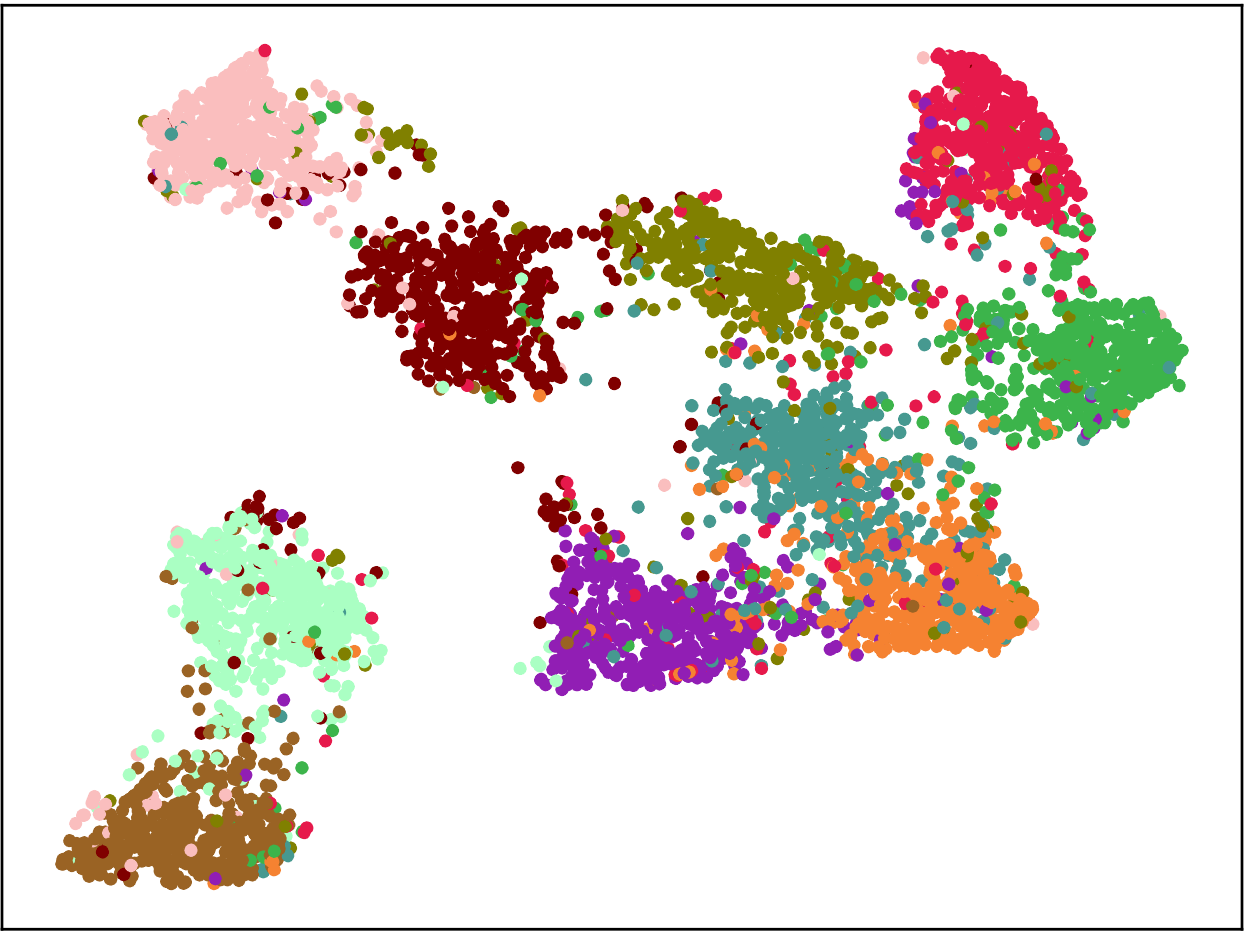}\tabularnewline
(a) & (b) & (c)\tabularnewline
\includegraphics[width=0.3275\columnwidth]{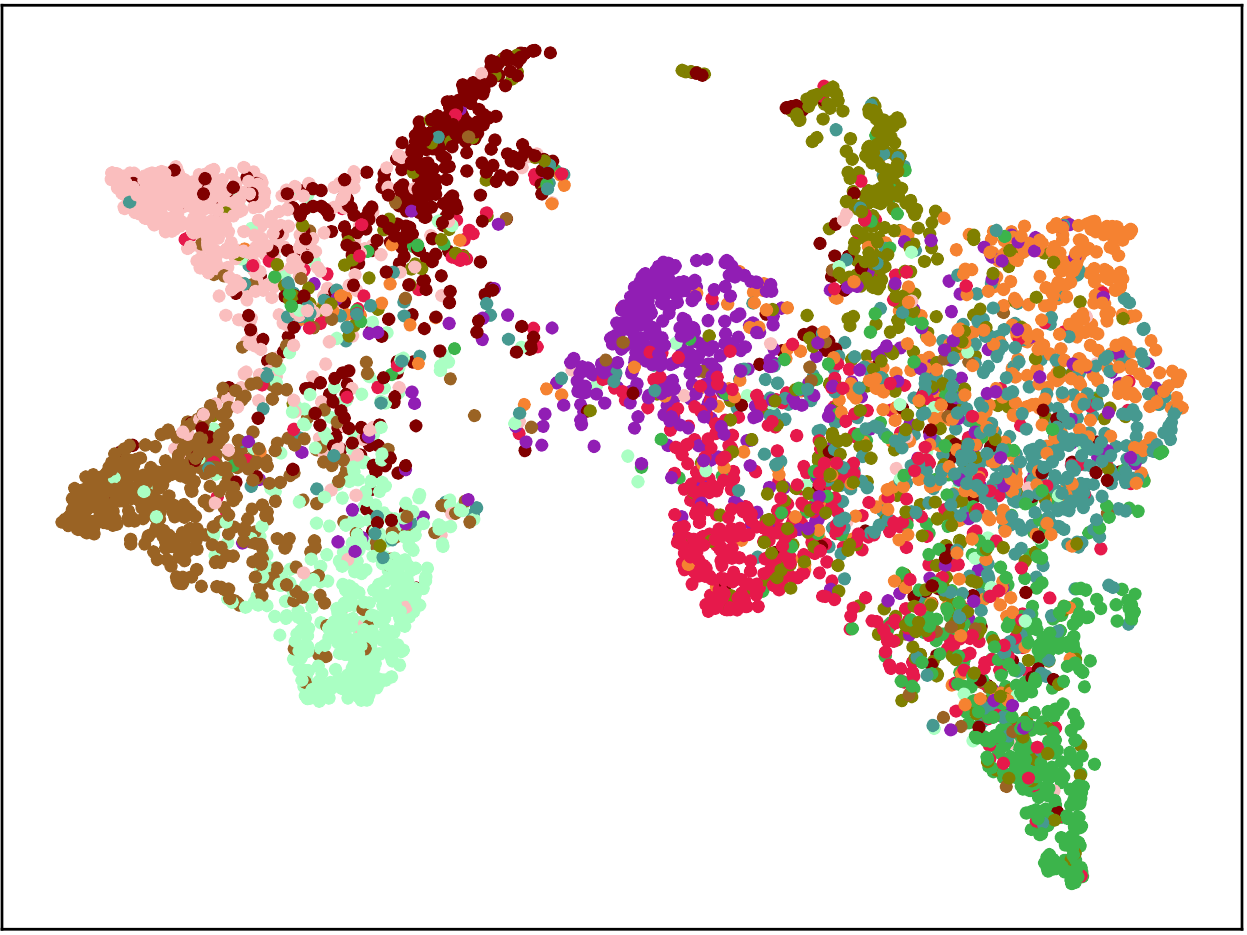} & \includegraphics[width=0.3275\columnwidth]{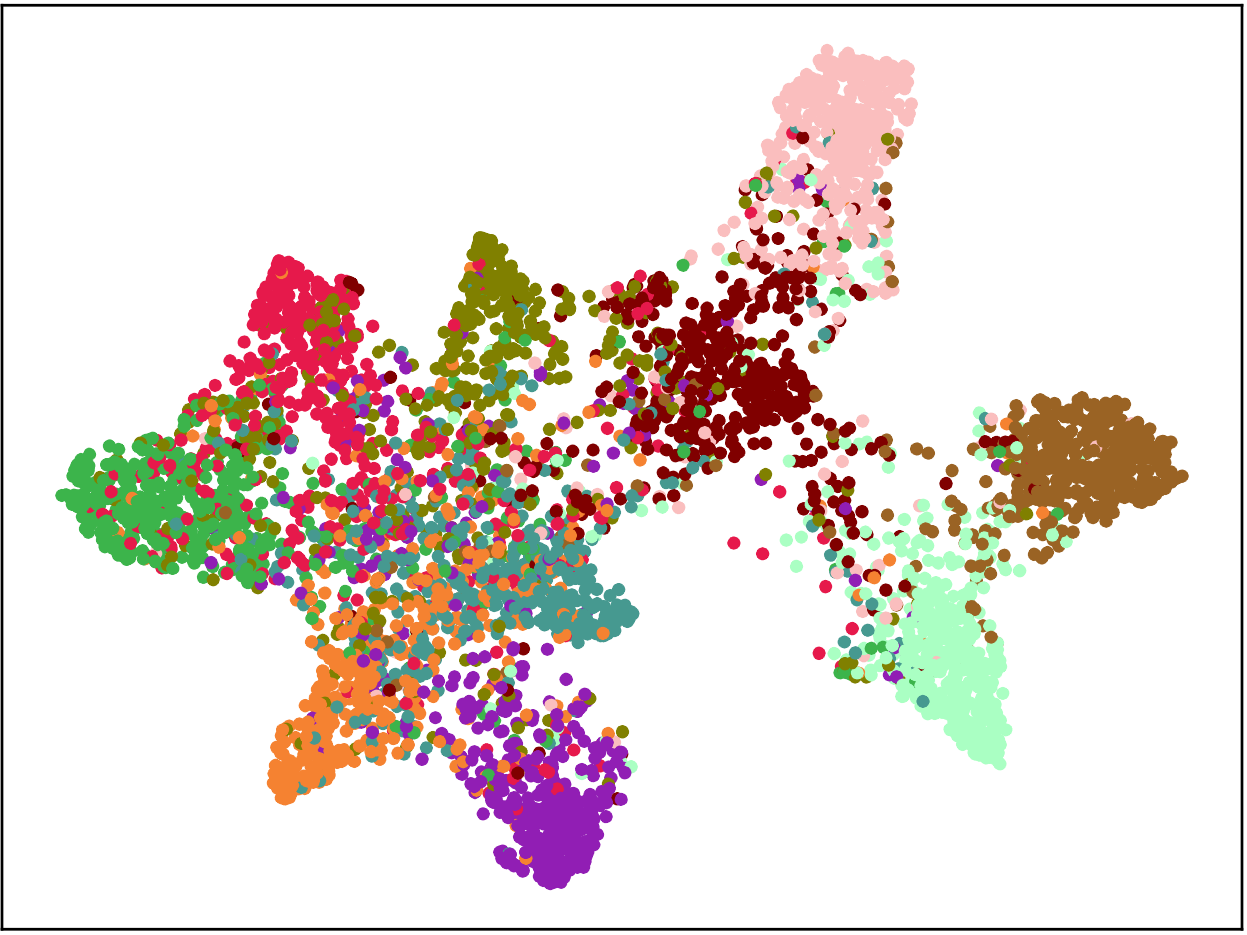} & \includegraphics[width=0.3275\columnwidth]{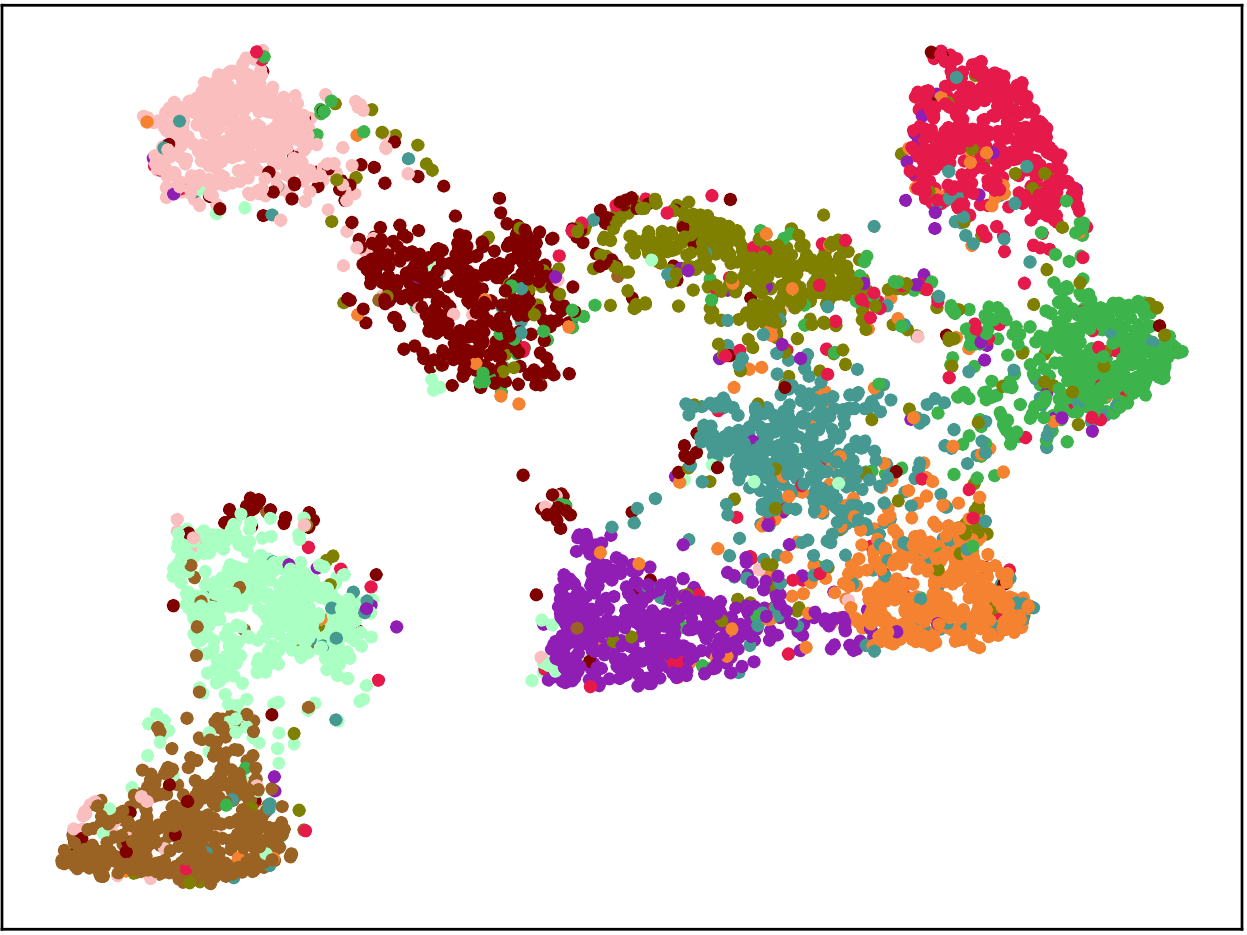}\tabularnewline
(d) & (e) & (f)\tabularnewline
\end{tabular}\caption{\label{fig:UMAP--embeddings}UMAP \cite{2018_JOSS_UMAP} embeddings
for training (top) with 80\% of label noise and validation (bottom) on CIFAR-10 with (a)(d) cross-entropy loss from Eq. \ref{eq:Cross-ent}, (b)(e) mixup \cite{2018_ICLR_mixup} and (c)(f) our proposed M-DYR-H.}
\vskip -0.2in 
\end{figure}

The proposed dynamic hard bootstrapping exhibits better performance
than the state-of-the-art static version~\cite{2015_ICLR_Bootstrapping}. 
It is, however, not better than the performance of \emph{mixup} data augmentation, which exhibits excellent robustness to label noise (M in Table~\ref{tab:JointPerformance}). The
fusion approach from Eq.~\eqref{eq:SMixDHardLoss} (M-DYR-H) and its
soft alternative (M-DYR-S), which combines the per-sample weighting
of dynamic bootstrapping and robustness to fitting noise labels of
\emph{mixup}, achieves a remarkable improvement in accuracy under high
noise levels. Table~\ref{tab:JointPerformance} reports
outstanding accuracy for 80\% of label noise, a case where we
improve upon \emph{mixup}~\cite{2018_ICLR_mixup} in best (last) accuracy
of 71.6 (46.7) in CIFAR-10 and 30.8 (17.6) in CIFAR-100 to 86.8 (86.6)
and 48.2 (47.2) using the hard alternative (M-DYR-H). It is important to highlight that we achieve quite similar best and
last performance for all levels of label noise in CIFAR datasets,
indicating that the proposed method is robust to varying noise levels. Figure \ref{fig:UMAP--embeddings} shows uniform manifold approximation and 
projection (UMAP) embeddings~\cite{2018_JOSS_UMAP} of the 512 features in the penultimate 
fully-connected layer of PreAct ResNet-18 trained using our method, and compares them with those found using cross-entropy and \emph{mixup}. 
The separation among classes appears visually more distinct using the proposed
objective.
\begin{table}[t]
\begin{centering}
\caption{\label{tab:ExtremeNoise}Validation accuracy on CIFAR-10 (top) and CIFAR-100
(bottom) with extreme label noise. Key: M (mixup), MD (dynamic mixup),
DYR (dynamic bootstrapping + reg. from Eq.~\eqref{eq:Regularization}),
H (hard), and SH (soft to hard). ({*}) denotes that we have run the
algorithm. Bold indicates best performance.}
\vskip 0.15in 
\par\end{centering}
\begin{centering}
\begin{small}%
\begin{tabular}{llrrrrr}
\toprule 
Alg./Noise level (\%) &  & 70 & 80 & 85 & 90\tabularnewline
\midrule 
\multirow{2}{*}{M-DYR-H} & Best & \textbf{89.6} & \textbf{86.8} & 71.6 & 40.8\tabularnewline
 & Last & \textbf{89.6} & \textbf{86.6} & 71.4 & 9.9\tabularnewline
\midrule 
\multirow{2}{*}{MD-DYR-H} & Best & 86.6 & 83.2 & \textbf{79.4} & 56.7\tabularnewline
 & Last & 85.2 & 80.5 & \textbf{77.3} & 50.0\tabularnewline
\midrule 
\multirow{2}{*}{MD-DYR-SH} & Best & 84.6 & 82.4 & 79.1 & \textbf{69.1}\tabularnewline
 & Last & 80.8 & 77.8 & 73.9 & \textbf{68.7}\tabularnewline
\bottomrule 
\toprule 
Alg./Noise level (\%) &  & 70 & 80 & 85 & 90\tabularnewline
\midrule 
\multirow{2}{*}{M-DYR-H} & Best & \textbf{54.4} & \textbf{48.2} & \textbf{29.9} & 12.5\tabularnewline
 & Last & \textbf{52.5} & \textbf{47.6} & \textbf{29.4} & 8.6\tabularnewline
\midrule 
\multirow{2}{*}{MD-DYR-H} & Best & 54.4 & 47.7 & 19.8 & 13.5\tabularnewline
 & Last & 50.8 & 41.7 & 8.3 & 3.9\tabularnewline
\midrule 
\multirow{2}{*}{MD-DYR-SH} & Best & 53.1 & 41.6 & 28.8 & \textbf{24.3}\tabularnewline
 & Last & 47.7 & 35.4 & 24.4 & \textbf{20.5}\tabularnewline
\bottomrule 
\end{tabular}\end{small}
\par\end{centering}
\centering{}\vskip -0.1in 
\end{table}

\subsection{On the limits of the proposed approach\label{subsec:On-the-limits}}

Table~\ref{tab:ExtremeNoise} explores convergence under extreme
label noise conditions, showing that the proposed approach M-DYR-H
fails to converge in CIFAR-10 with 90\% label noise.
Here we propose minor modifications to achieve convergence.

When clean and noisy samples
are combined by \emph{mixup} they are given the same importance of approximately
$\delta=0.5$ (as $\alpha=\beta=32$). While noisy samples
benefit from mixing with clean ones, clean samples are contaminated by 
noisy ones, whose training objective is incorrectly modified.
We propose a dynamic \emph{mixup} strategy in the input
that uses a different $\delta$ for each sample to reduce the contribution
of noisy samples when they are mixed with clean ones: 
\begin{equation}
x=\left(\frac{\delta_{p}}{\delta_{p}+\delta_{q}}\right)x_{p}+\left(\frac{\delta_{q}}{\delta_{p}+\delta_{q}}\right)x_{q},
\end{equation}
where $\delta_{p}=p\!\left(k=0\mid\ell_{p}\right)$ and $\delta_{q}=p\!\left(k=0\mid\ell_{q}\right)$,
i.e. we use the noise probability from our BMM to guide \emph{mixup}
in the input. Note that for clean-clean and noisy-noisy cases, the behavior remains similar to \emph{mixup} with $\alpha=\beta=32$,
which leads to $\delta\approx0.5$ (i.e. $\delta_{p}\approx\delta_{q}\Rightarrow$
$\delta_{p}/(\delta_{p}+\delta_{q})\approx0.5$).
This configuration simplifies the input to the network when mixing a sample whose label is potentially useless, 
while retaining the strengths of \emph{mixup} for clean-clean and 
noisy-noisy combinations. This is used with the original \emph{mixup} strategy (Eq.~\eqref{eq:SMixDHardLoss})
to benefit from the regularization that an additional
label provides. Table~\ref{tab:ExtremeNoise} presents the results
of this approach (MD-DYR-H), which exhibits more stable convergence for 90\% 
label noise in both datasets.

Table~\ref{tab:JointPerformance} reported that hard bootstrapping
works better than the soft alternative. Unfortunately, hard
bootstrapping under high levels of label noise causes large variations
in the loss that lead to drops in performance. To ameliorate such instabilities,
we propose a decreasing softmax technique \cite{2005_ECML_DecreasingSoftMax}
to progressively move from a soft to a hard dynamic bootstrapping.
This is implemented by modifying the softmax temperature $T$
in: 
\begin{equation}
h_{ij}=\frac{\exp\!\left(s_{ij}/T\right)}{\sum_{k=1}^{N}\exp\!\left(s_{ik}/T\right)},
\end{equation}
where $s_{ij}$ denotes the score obtained
in the last layer of the CNN model class $j$ of sample $x_{i}$. By default
$T=1$ gives the soft alternative of Eq.~\eqref{eq:SMixDHardLoss}.
To move from soft to hard bootstrapping we linearly
reduce the temperature for $h_{p}$
and $h_{q}$ until we reach a final temperature in a certain
epoch ($T=0.001$ and epoch 200 in our experiments). We experimented
with linear, logarithmic, tanh, and step-down temperature decays with
similar results. This decreasing softmax MD-DYR-SH obtains
much improved accuracy for 90\% of label noise (69.1 for CIFAR-10 and
24.3 for CIFAR-100), while slightly decreasing accuracy compared
to M-DYR-H and MD-DYR-H at lower noise levels. Note that we significantly
outperform the best state-of-the-art we are aware for 90\% of label
noise, which is 58.3\% and 58.0\% for best and last validation accuracies (reported in \cite{2018_CVPR_JointOpt} with a PreAct ResNet-32 on CIFAR-10). The training process is slightly modified to introduce dynamic \emph{mixup} (epoch 106) before  bootstrapping (epoch 111) for MD-DYR-H and MD-DYR-SH.
\begin{table}[t]
\begin{centering}
\caption{\label{tab:ComparisonSoA1}Comparison with the state-of-the-art
in terms of validation accuracy on CIFAR-10 (top) and CIFAR-100 (bottom).
Key: M (mixup), MD (dynamic mixup), DYR (dynamic bootstrapping + reg.
from Eq. \ref{eq:Regularization}), H (hard) and SH (soft to hard).
({*}) denotes that we have run the algorithm. Bold indicates best
performance.}
\vskip 0.15in 
\par\end{centering}
\begin{centering}
\begin{small}\resizebox{1\columnwidth}{!}{%
\begin{tabular}{llrrrrr}
\toprule 
Alg./Noise level (\%) &  & 0 & 20 & 50 & 80 & 90\tabularnewline
\midrule 
\multirow{2}{*}{\cite{2015_ICLR_Bootstrapping}{*}} & Best & 94.7 & 86.8 & 79.8 & 63.3 & 42.9\tabularnewline
 & Last & 94.6 & 82.9 & 58.4 & 26.8 & 17.0\tabularnewline
\midrule 
\multirow{2}{*}{\cite{2017_CVPR_ForwardLoss}{*}} & Best & 94.7 & 86.8 & 79.8 & 63.3 & 42.9\tabularnewline
 & Last & 94.6 & 83.1 & 59.4 & 26.2 & 18.8\tabularnewline
\midrule 
\multirow{2}{*}{\cite{2018_ICLR_mixup}{*}} & Best & \textbf{95.3} & \textbf{95.6} & 87.1 & 71.6 & 52.2\tabularnewline
 & Last & \textbf{95.2} & 92.3 & 77.6 & 46.7 & 43.9\tabularnewline
\midrule 
\multirow{2}{*}{M-DYR-H} & Best & 93.6 & 94.0 & \textbf{92.0} & \textbf{86.8} & 40.8\tabularnewline
 & Last & 93.4 & \textbf{93.8} & \textbf{91.9} & \textbf{86.6} & 9.9\tabularnewline
\midrule 
\multirow{2}{*}{MD-DYR-SH} & Best & 93.6 & 93.8 & 90.6 & 82.4 & \textbf{69.1}\tabularnewline
 & Last & 92.7 & 93.6 & 90.3 & 77.8 & \textbf{68.7}\tabularnewline
\bottomrule 
\toprule 
Alg./Noise level (\%) &  & 0 & 20 & 50 & 80 & 90\tabularnewline
\midrule 
\multirow{2}{*}{\cite{2015_ICLR_Bootstrapping}{*}} & Best & \textbf{76.1} & 62.1 & 46.6 & 19.9 & 10.2\tabularnewline
 & Last & \textbf{75.9} & 62.0 & 37.9 & 8.9 & 3.8\tabularnewline
\midrule 
\multirow{2}{*}{\cite{2017_CVPR_ForwardLoss}{*}} & Best & 75.4 & 61.5 & 46.6 & 19.9 & 10.2\tabularnewline
 & Last & 75.2 & 61.4 & 37.3 & 9.0 & 3.4\tabularnewline
\midrule 
\multirow{2}{*}{\cite{2018_ICLR_mixup}{*}} & Best & 74.8 & 67.8 & 57.3 & 30.8 & 14.6\tabularnewline
 & Last & 74.4 & 66.0 & 46.6 & 17.6 & 8.1\tabularnewline
\midrule 
\multirow{2}{*}{M-DYR-H} & Best & 70.3 & 68.7 & 61.7 & \textbf{48.2} & 12.5\tabularnewline
 & Last & 66.2 & 68.5 & 58.8 & \textbf{47.6} & 8.6\tabularnewline
\midrule 
\multirow{2}{*}{MD-DYR-SH} & Best & 73.3 & \textbf{73.9} & \textbf{66.1} & 41.6 & \textbf{24.3}\tabularnewline
 & Last & 71.3 & \textbf{73.4} & \textbf{65.4} & 35.4 & \textbf{20.5}\tabularnewline
\bottomrule 
\end{tabular}}\end{small}
\par\end{centering}
\centering{}\vskip -0.1in 
\end{table}
\begin{table}[t]
\begin{centering}
\caption{\label{tab:ComparisonSoA2}Comparison with the state-of-the-art
in terms of validation accuracy on CIFAR-10 (top) and CIFAR-100 (bottom).
Key: M (mixup), MD (dynamic mixup), DYR (dynamic bootstrapping + reg.
from Eq. \ref{eq:Regularization}), H (hard), SH (soft to hard), WRN
(Wide ResNet), PRN (PreActivation ResNet, and GCNN (Generic CNN). Bold
indicates best performance.}
\vskip 0.15in 
\par\end{centering}
\begin{centering}
\begin{small}\resizebox{1\columnwidth}{!}{%
\begin{tabular}{llrrrr}
\toprule 
\multirow{2}{*}{Algorithm} & \multirow{2}{*}{Architecture} & \multicolumn{4}{c}{Noise level (\%)}\tabularnewline
 &  & 20 & 40 & 60 & 80\tabularnewline
\midrule 
\cite{2018_ICML_MentorNet} & WRN-101 & 92.0 & 89.0 & - & 49.0\tabularnewline
\cite{2018_ICML_DimDriven} & GCNN-12 & 85.1 & 83.4 & 72.8 & -\tabularnewline
\cite{2018_ICML_L2ReweightNoise} & WRN-28 & - & 86.9 & - & -\tabularnewline
\cite{2018_CVPR_IterativeNoise} & GCNN-7 & 81.4 & 78.2 & - & -\tabularnewline
M-DYR-H & PRN-18 & \textbf{94.0} & \textbf{92.8} & \textbf{90.3} & 46.3\tabularnewline
MD-DYR-SH & PRN-18 & 93.8 & 92.3 & 86.1 & \textbf{74.1}\tabularnewline
\bottomrule 
\toprule 
\multirow{2}{*}{Algorithm} & \multirow{2}{*}{Architecture} & \multicolumn{4}{c}{Noise level (\%)}\tabularnewline
 &  & 20 & 40 & 60 & 80\tabularnewline
\midrule 
\cite{2018_ICML_MentorNet} & WRN-101 & 73.0 & 68.0 & - & 35.0\tabularnewline
\cite{2018_ICML_DimDriven} & RN-44 & 62.2 & 52.0 & 42.3 & -\tabularnewline
\cite{2018_ICML_L2ReweightNoise} & WRN-28 & - & 61.3 & - & -\tabularnewline
M-DYR-H & PRN-18 & 70.0 & 64.4 & 58.1 & \textbf{45.5}\tabularnewline
MD-DYR-SH & PRN-18 & \textbf{73.7} & \textbf{70.1} & \textbf{59.5} & 39.5\tabularnewline
\bottomrule 
\end{tabular}}\end{small}
\par\end{centering}
\centering{}\vskip -0.1in 
\end{table}

\subsection{Comparison with related approaches\label{subsec:Comparison-against-related}}

Table~\ref{tab:ComparisonSoA1} compares with related works for different levels of label noise using a common architecture and the 300 epochs training scheme (see
Subsection \ref{subsec:Datasets-and-implementation}) .
We introduce bootstrapping in epoch 105 for \cite{2015_ICLR_Bootstrapping}
for the proposed methods, estimate the $T$ matrix of \cite{2017_CVPR_ForwardLoss}
in epoch 75 (as done in \cite{2018_NIPS_GoldLoss}), and use the configuration reported in \cite{2018_ICLR_mixup} for \emph{mixup}. We outperform the
related work in the presence of label noise, obtaining remarkable improvements
for high levels of noise (80\% and 90\%) where the compared
approaches do not learn as well from the noisy samples (see best accuracy)
and do not prevent fitting noisy labels (see last accuracy).

As noted in Subsection \ref{subsec:Datasets-and-implementation},
when introducing label noise the true label can be excluded from the
candidates. In this case label noise is defined as the percentage of incorrect labels
instead of random ones (i.e. the criterion followed in previous experiments), a criterion adopted by several other authors \cite{2018_ICML_MentorNet,2018_ICML_DimDriven,2018_ICML_L2ReweightNoise,2018_CVPR_IterativeNoise}. We also run our proposed approach under this setup to allow  quantitative comparison (Table~\ref{tab:ComparisonSoA2}). The proposed method outperforms all related work in CIFAR-10 and CIFAR-100 with MD-DYR-SH,
while the results for M-DYR-H are slightly below those of~\cite{2018_ICML_MentorNet}
for low label noise levels in CIFAR-100. Nevertheless, these results should be interpreted with care due to the different architectures employed and
the use of sets of clean data during training in \cite{2018_ICML_MentorNet}
and \cite{2018_ICML_L2ReweightNoise}.

\subsection{Generalization of the proposed approach\label{subsec:Additional datasets}}

\begin{table}[t]
\begin{centering}
\caption{\label{tab:TinyImageNet}Comparison of test accuracy on TinyImageNet. Key: M (mixup) , DYR (dynamic bootstrapping + reg. from
Eq. \ref{eq:Regularization}), H (hard), and SH (soft to hard). ({*})
denotes that we have run the algorithm. Bold indicates best performance.}
\vskip 0.15in 
\par\end{centering}
\begin{centering}
\begin{small}%
\begin{tabular}{llrrr}
\toprule 
Alg./Noise level (\%) &  & 20 & 50 & 80\tabularnewline
\midrule 
\multirow{2}{*}{\cite{2018_ICLR_mixup}{*}} & Best & 53.2 & 41.7 & 18.9\tabularnewline
 & Last &  49.4 & 31.1 & 8.7\tabularnewline
\midrule 
\multirow{2}{*}{M-DYR-H} & Best & 51.8 & 44.4 & 18.3\tabularnewline
 & Last &  51.6 &  43.6 &  17.7\tabularnewline
\midrule 
\multirow{2}{*}{MD-DYR-SH} & Best & \textbf{60.0} & \textbf{50.4} & \textbf{24.4}\tabularnewline
 & Last & \textbf{ 59.8} & \textbf{ 50.0} & \textbf{ 19.6}\tabularnewline
\bottomrule 
\end{tabular}\end{small}
\par\end{centering}
\centering{}\vskip -0.1in 
\end{table}
Table~\ref{tab:TinyImageNet} shows the results of the proposed
approaches M-DYR-H and MD-DYR-SH compared to \emph{mixup} \cite{2018_ICLR_mixup}
on TinyImageNet to demonstrate that our approach is useful far from
CIFAR data. The proposed approach clearly outperforms \cite{2018_ICLR_mixup}
for different levels of label noise, obtaining consistent results
with the CIFAR experiments. Note that we use
the same network, hyperparameters, and learning rate policy as with CIFAR.
Furthermore, we tested our approach in real-world label noise
by evaluating our method on Clothing1M \cite{2015_CVPR_GraphModelNoise},
which contains non-uniform label noise with label flips concentrated
in classes sharing similar visual patterns with the true class. We
followed a similar network and procedure as \cite{2018_CVPR_JointOpt}
with ImageNet pre-trained weights and ResNet-50, obtaining over 71\%
test accuracy, which falls short of the state-of-the-art (72.23\%  \cite{2018_CVPR_JointOpt}). We found
that finetuning a pre-trained network for one epoch, as done in \cite{2018_CVPR_JointOpt},
easily fits label noise limiting our unsupervised label noise model. We believe this occurs
due to the structured noise and the small learning rate. Training with cross-entropy alone gives test accuracy over 69\%, suggesting that the configurations used might be suboptimal.

\section{Conclusions}

This paper presented a novel approach on training under label noise
with CNNs that does not require any set of clean data. We proposed
to fit a beta mixture model to the cross-entropy loss of each sample
and model label noise in an unsupervised way. This model is used
to implement a dynamic bootstrapping loss that relies either on the
network prediction or the ground-truth (and potentially noisy) labels
depending on the mixture model. We combined this dynamic bootstrapping
with mixup data augmentation to implement an incredibly robust loss
correction approach. We conducted extensive experiments on CIFAR-10
and CIFAR-100 to show the strengths and weaknesses of our approach
demonstrating outstanding performance. We further proposed to use
our beta mixture model to guide the combination of \emph{mixup} data augmentation
to assure reliable convergence under extreme noise levels. The approach generalizes well to TinyImageNet but shows some limitations under non-uniform noise in Clothing1M that we will explore in future research.

\section*{Acknowledgements}
This work was supported by Science Foundation Ireland (SFI) under grant numbers SFI/15/SIRG/3283 and SFI/12/RC/2289.

\bibliography{refs}
 \bibliographystyle{icml2019}
 

\makeatother

\twocolumn[
\icmltitle{Supplementary material for the paper ``Unsupervised Label Noise Modeling and Loss Correction''}
]

\appendix
\icmltitlerunning{Unsupervised label noise modeling and loss correction: Supplementary material}
\section{Beta Mixture Model (BMM)}

This section extends the discussion of the proposed unsupervised BMM in the main paper providing detail on several more aspects.

\paragraph{BMM performance under low levels of label noise} We seek robust representation learning in the presence of label noise,
which may occur when images are automatically labeled. Performance
will likely drop in carefully annotated datasets with near 0\% noise
because the loss distribution is not a two-component mixture. In this
situation the BMM classifies almost all samples as clean, but some
estimation errors may occur, which lead to a reliance on the sometimes
incorrect network prediction instead of the true clean label. Nevertheless,
for 20\% noise, we outperform the compared state-of-the-art at the
end of the training, demonstrating improved robustness for low noise
levels.

\begin{figure}[!t]
\centering{}%
\begin{tabular}{c}
\includegraphics[width=0.95\columnwidth]{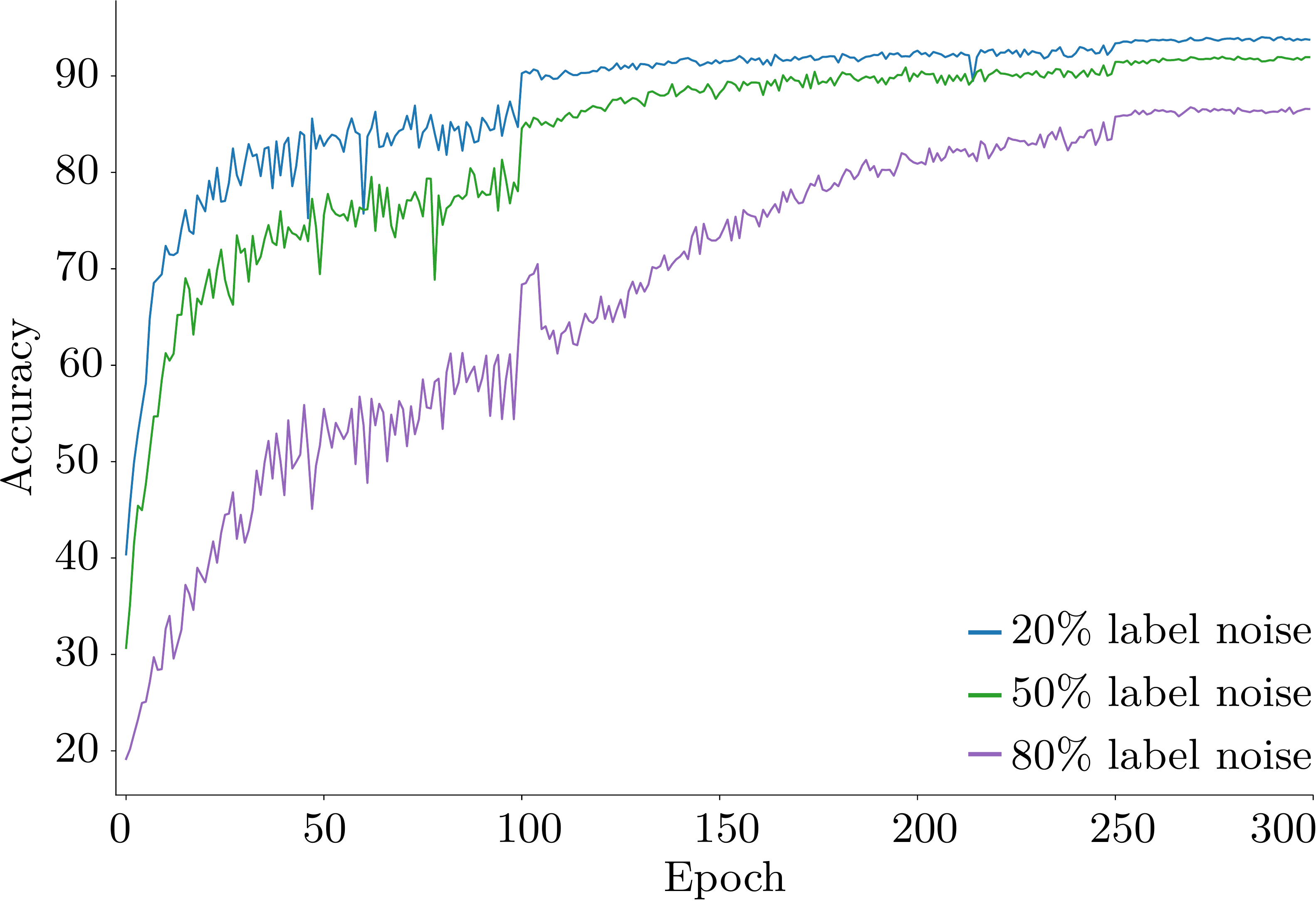}\tabularnewline
(a)\tabularnewline
\includegraphics[width=0.95\columnwidth]{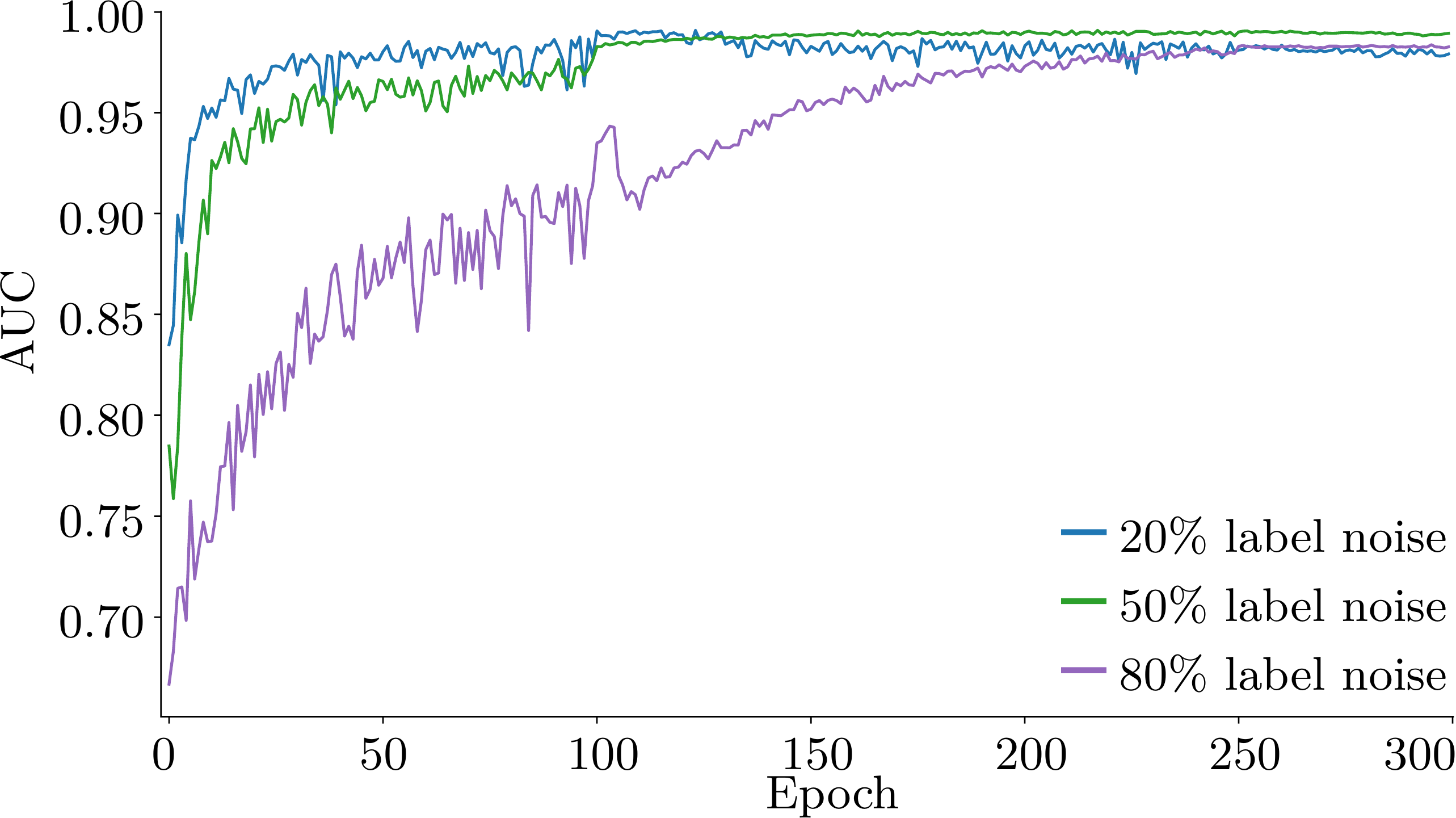}\tabularnewline
(b)\tabularnewline
\includegraphics[width=0.95\columnwidth]{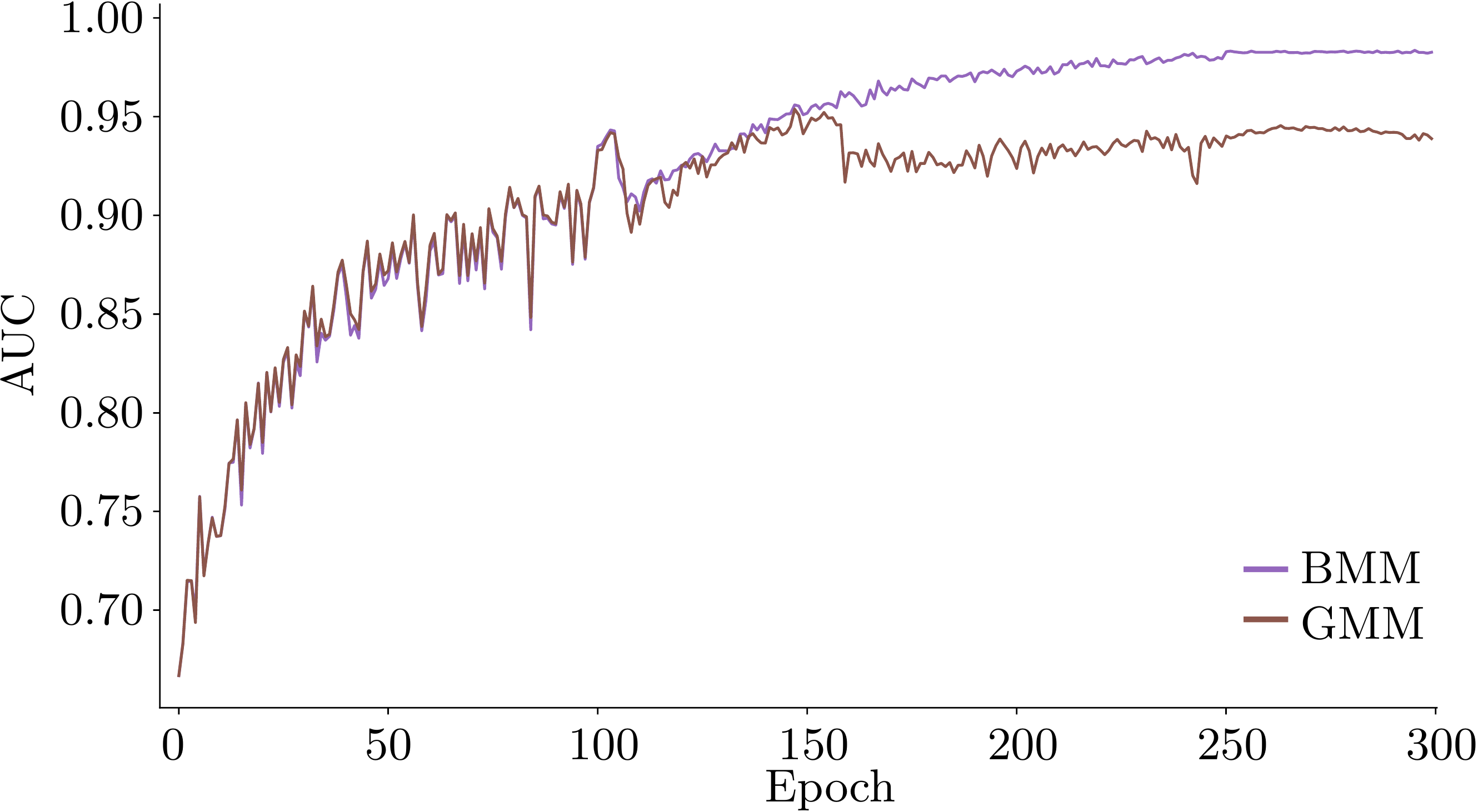}\tabularnewline
(c)\tabularnewline
\end{tabular}\caption{\label{fig:Proposed-approach-M-DYR-H}M-DYR-H results
on CIFAR-10 for (a) image classification and (b) clean/noisy classification
of the BMM. (c) comparison of GMM and BMM for clean/noisy
classification with 80\% label noise.}
\end{figure}

\paragraph{BMM parameter estimation frequency} The BMM parameters are re-estimated after every epoch once the loss
correction begins (i.e. there is an initial warm-up as noted in Subsection
4.1 with no loss correction) by computing the cross-entropy loss from
a forward pass with the original (potentially noisy) labels. We
also tested our approach M-DYR-H (CIFAR-10, 80\% of label noise) changing
the estimation period to 5 and 0.5 epochs, observing no decrease in
accuracy. While the original configuration presented in Figure \ref{fig:Proposed-approach-M-DYR-H}(a)
reaches 86.8 (86.6) for best (last), every 5 epochs leads to (86.9)
86.8 and every 0.5 to 88.0 (87.5).

\paragraph{BMM classification accuracy and robustness} Figure \ref{fig:Proposed-approach-M-DYR-H}(b) shows
the clean/noisy classification capabilities of the
BMM in terms of Area Under the Curve (AUC) evolution during training,
demonstrating that performance and robustness are consistent across
noise levels. In particular, the experiment on CIFAR-10
with M-DYR-H exceeds 0.98 AUC for 20, 50 and 80\% label noise.
AUC increases during training and increases faster
for lower noise levels, showing increasingly better clean/noisy
discrimination related to consistent BMM predictions over time. 

\paragraph{Effect of BMM classification accuracy on image classification accuracy}
BMM prediction accuracy is essential for high image classification accuracy, as demonstrated by the tendency for both image classification and BMM accuracy to increase together in Figure~\ref{fig:Proposed-approach-M-DYR-H}(a) and (b), especially for higher noise levels. 
Figure~\ref{fig:Proposed-approach-M-DYR-H}(c) further verifies this relationship by comparing the BMM with a GMM (Gaussian Mixture Model) on CIFAR-10 with M-DYR-H and 80\% label noise. 
The GMM gives both less accurate clean/noisy discrimination and worse image classification results (clean/noisy AUC drops from 0.98 to 0.94, while image classification accuracy drops from 86.6 to 83.5).


\paragraph{Performance attributable to the BMM}
Incorporating the BMM results in a loss that goes beyond mere regularization. This can be verified by removing the BMM
and assigning fixed weights in the bootstrapping loss (0.8 to GT and
0.2 to network prediction, keeping mixup for robustness). This leads
to a drop from 86.6 for M-DYR-H to 74.6 in the last epoch (80\% of
label noise on CIFAR10).


\section{Hyperparameters}

We stress that experiments across all datasets share the same hyperparameter
configuration and lead to consistent improvements over the state-of-the-art,
demonstrating that the general approach does not require carefully
tuned hyperparams. Indeed, we are likely reporting suboptimal results
that could be improved with a label noise free validation set, though
availability of this set is not assumed in this paper.

Starting training with high learning rates is important:
training more epochs leads to better performance, as mixup together
with a high learning rate helps prevent fitting label noise. This warm-up
learns the structured data (mainly associated to clean samples) and
helps separate the losses between clean/noisy samples for a better
BMM fit.

\paragraph{Experiment details} All experiments used the following setup and hyperparameter configuration:
\begin{description}
\item[Preprocessing] Images are normalized and augmented by random horizontal flipping. We use 32\texttimes 32 random crops after zero padding with 4 pixels on each side.
\item[Network] A PreAct ResNet-18 is trained from scratch using PyTorch 0.4.1. Default PyTorch initialization is used on all layers. 
\item[Optimizer] SGD with momentum (0.9), weight decay of $10^{-4}$, and batch size 128.
\item[Training schedule without mixup] Training for 120 epochs in total. We reduce the initial learning rate (0.1) by a factor of 10 after 30, 80, and
110 epochs. Warm-up for 30 epochs, i.e. bootstrapping (when used) starts
in epoch 31. This configuration is used in all experiments in Table
1.
\item[Training schedule with mixup] Training for 300 epochs in total.
We reduce the initial learning rate  (0.1) by a factor of 10 after 100 and 250
epochs. Warm-up for 105 epochs, i.e. bootstrapping starts
in epoch 106 when used (note: the warmup period can be much longer when using mixup because it mitigates fitting label noise. Mixup $\alpha = 32$. This configuration is used for all experiments \textit{excluding}
those in Table 1.
\item 
\end{description}
Regarding BMM parameter estimation: parameters are fit automatically using 10 EM iterations as noted in the paper. 
We also ran M-DYR-H (80\% of label noise, CIFAR-10) using 5 and 20 EM iterations, obtaining 87.4 (87.2)
and 86.9 (86.3) for best (last) epoch, suggesting that the method is relatively robust to this hyperparameter.

\end{document}